\documentclass[10pt,twocolumn,letterpaper]{article}

\usepackage[final]{cvpr}      
\pdfoutput=1
\usepackage[mathscr]{eucal}
\usepackage[table]{xcolor}

\usepackage[pagebackref,breaklinks,colorlinks]{hyperref}
\usepackage[accsupp]{axessibility}
\usepackage{CJKutf8}
\usepackage{graphicx}
\usepackage{float}
\usepackage{array}
\usepackage{amsmath,amssymb,amsfonts}
\usepackage{booktabs}
\usepackage{tabularx}
\usepackage{multirow}
\usepackage{wrapfig}
\usepackage{amssymb}
\usepackage{paralist} 

\usepackage[accsupp]{axessibility}

\usepackage[capitalize]{cleveref}
\crefname{section}{Section}{Secs.}
\Crefname{section}{Section}{Sections}
\Crefname{table}{Table}{Tables}
\crefname{table}{Table}{Tabs.}

\def\cvprPaperID{*****} 
\def\confName{CVPR}
\def\confYear{2023}

\begin{document}

\title{Semantic Segmentation on VSPW Dataset through Contrastive Loss and Multi-dataset Training Approach}

\author{Min Yan$^{1 \dag}$ ~Qianxiong Ning$^{1,2 \dag}$ ~Qian Wang$^{1}$ \\
$^{1}$ China Mobile Research Institute, \\
$^{2}$ Xi'an Jiaotong University \\
{\tt\small yanminai@chinamobile.com; n1262986973@stu.xjtu.edu.cn; wangqianrg@chinamobile.com}}

\maketitle

\footnotetext{$^\dag$Equal contribution.}

\begin{abstract}
Video scene parsing incorporates temporal information, which can enhance the consistency and accuracy of predictions compared to image scene parsing. The added temporal dimension enables a more comprehensive understanding of the scene, leading to more reliable results. This paper presents the winning solution of the CVPR2023 workshop for video semantic segmentation, focusing on enhancing Spatial-Temporal correlations with contrastive loss. In addition, we explore the influence of multi-dataset training by utilizing a label-mapping technique. And the final result is aggregating the output of the above two models. 
Our approach achieves 65.95\% mIoU performance on the VSPW dataset, ranked 1st place on the VSPW challenge at CVPR 2023.
\end{abstract}

\section{Introduction}
\label{sec:intro}
The Video Scene Parsing in the Wild (VSPW)~\cite{miao2021vspw}~is a recently introduced video semantic segmentation dataset comprising 3536 videos with an average duration of approximately 5 seconds and a frame rate of 15. This dataset encompasses annotations for 124 categories. The main objective of the challenge is to perform semantic segmentation of the test set videos in the VSPW dataset by assigning predefined semantic labels to pixels across all frames. The prominent evaluation metric for the challenge is the mean Intersection over Union (mIoU).

After conducting a thorough investigation of the dataset, we made numerous attempts during the competition, and we will describe our experimental process in the subsequent section to inspire future researchers. We first compared several open-source models, considering both model size and performance. As a result, we selected the existing open-source model with decent performance on the ADE20k\cite{zhou2017scene} dataset, which is Vit-Adapter + Mask2former\cite{chen2022vision,cheng2021mask2former}, as our baseline model.

Furthermore, we recognized that Transformer-based models have powerful learning capabilities, and additional data training can improve the model's learning ability. Therefore, during the training phase, we adopted the joint training method on multiple datasets, which significantly affected the training dataset. However, due to the distribution differences between the validation and test sets, introducing the ADE20k dataset did not achieve the desired results. In the final outcome, we only used the coco-stuff 160k\cite{caesar2018coco} and VSPW datasets for joint training. Specifically, when processing the  COCO-Stuff dataset\cite{caesar2018coco}, we used the label-mapping method to map the semantic labels of the coco-stuff training and validation sets to the categories in the VSPW dataset with the closest semantic meaning. Then, we set a threshold to retain images whose effective labels were greater than 80\%  as the final extra-training set data.

To enhance the correlation between temporal and spatial features in video semantic segmentation, we considered that the predicted results should have strong frame-to-frame correlation, meaning that for the same object, the predicted results in adjacent frames of the same video should be as consistent as possible, while the semantic segmentation predicted results between different object categories should be as different as possible. Therefore, we introduced a contrastive loss function\cite{chen2020improved}. Specifically, when reading the data, we selected two adjacent frames as input for the training set. We calculated the difference between the semantic pixel features of different categories using the contrastive loss function to increase their differences while reducing the differences within the same category.

In the end, we combined the above two models for the model aggregation in the testing phase and achieved a mIoU of 65.95\% in the final test set.
\section{Method}{
\subsection{Baseline Model}{
Transformer-based models have shown excellent performance in semantic segmentation tasks due to their powerful learning capabilities. Therefore, we utilize Vit-Adapter and Mask2former model, a transformer-based model with a strong performance on the Ade20k leaderboard, as our baseline model. The Beitv2\cite{peng2022beit} is chosen as the backbone of our baselines, which is pre-trained on ImageNet21k\cite{deng2009imagenet}. The entire model has been fine-tuned on the COCO-Stuff and ADE20k datasets. 





\subsection{Multi-dataset training}
\label{subsec: Multi-dataset training}
{
To enhance the performance of our model, we have included additional training data beyond the VSPW training set, namely ADE20k and COCO datasets.
We attempted to train our model on multiple datasets through a label mapping technique where we processed the COCO-Stuff and ADE20k datasets by associating their labels with those of the VSPW dataset that had similar semantics. After processing, we filtered the resulting dataset to remove images with a lot of irrelevant pixels, resulting in a joint dataset. We regard the two datasets as a whole to train the network. However, in our final experimental results, we observed a significant improvement in performance using the COCO dataset, while ADE20k did not. We believe that the reason for the poor performance may be the significant differences in data distribution and scale between ADE20k and VSPW datasets.  Therefore, we did not use the ADE20k dataset for joint training in the final results.
}
\subsection{Contrastive Loss}{
We argue that pixel features of the same semantic class should be as consistent as possible for consecutive frames in a video sequence. In contrast, those of different semantic classes should be as distinct as possible. To establish temporal connections between the sequence features, we introduced a spatial-temporal contrastive loss function to establish continuity between adjacent video frames and increase the contrast between different semantic classes within the same image. Formally, our spatial-temporal contrastive loss is defined as:
\small
\begin{equation}
\mathcal{L}_{i} = \frac{1}{| \phi|}\sum_{x_i^+\in \phi}
\log\frac{\exp\left((x_i \cdot x_i^+)/\tau\right)}{\exp\left((x_i \cdot x_i^+)/\tau\right) + \sum_{j=1}^{K}\exp\left(x_i \cdot x_j^-)/\tau\right)}
\end{equation}
\begin{equation}
\mathcal{L}_{Nce} = -\frac{1}{M}\sum_{i=1}^{M}\mathcal{L}_{i}
\end{equation}
where $x_i$ is the input sample, including two consecutive frames from the encoder, The positive sample $x_i^+$ contains semantic labels that belong to the same category as $x_i$, while the negative sample $x_j^-$ contains semantic labels that belong to a different category. The temperature coefficient $\tau$ is used to control the smoothness of the probability distribution. $\phi$ denotes the positive samples for pixel $i$.The parameters $K$ denote the number of negative samples, respectively. The variable $M$ represents the number of patches obtained from the two-frame images after passing through the backbone and encoder in the NCE loss.

By minimizing this loss, we can establish spatial-temporal consistency between adjacent frames. However, we can only conduct experiments on resolution-cropped images with a crop size 480x480 due to memory limitations.
\subsection{Model aggregation}
 Aggregating models is a widely recognized technique for improving model performance. Training two models under different conditions can produce different results. Consequently, a combination of the outputs of these models through a weighted summation approach can yield a substantial improvement in the test sets of VSPW. This technique leverages the complementary strengths of each constituent model, resulting in a more robust and accurate prediction. The weighted summation approach is a simple and effective means of model aggregation, allowing for the integration of multiple models with minimal computational overhead. The specific results and other attempts will be described in detail in section~\ref{sec: experiments result}.
 \subsection{Loss function}
We train the  model with the following loss function
\begin{equation}
    \mathcal{L} = \lambda_{1}\mathcal{L}_{seg} + \lambda_{2}\mathcal{L}_{Nce}
\end{equation}
\begin{equation}
    \mathcal{L}_{seg} = \lambda_{3}\mathcal{L}_{Dice} + \lambda_{4}\mathcal{L}_{CE}
\end{equation}
while $\mathcal{L}_{seg}$ is consist of Dice loss\cite{li2019dice} and cross-entropy loss and $\lambda_{1}$, $\lambda_{2}$, $\lambda_{3}$, $\lambda_{4}$ is set to 1, 0.1, 5, 1 respectively.  We only used $\mathcal{L}_{seg}$ separately during multi-dataset joint training. After adding the contrastive loss, we used the above-mentioned loss $\mathcal{L}$ on the model.

\section{Experiments Result}.
\label{sec: experiments result}
\subsection{Dataset and Evaluation Metrics}
VSPW dataset is a large-scale video semantic segmentation dataset. It includes 3536 videos and 251633 frames with 124 classes. The training, validation, and testing sets contain 2806/343/387 videos with 198244,24502,28887 frames, respectively.   

COCO-Stuff dataset is an image semantic segmentation dataset that contains about 123k images with 182 classes. We include some of the COCO-Stuff dataset in training. The usage of this dataset can be found in section~\ref{subsec: Multi-dataset training}

We adopt mean Interaction over Union(mIoU) as evaluation metrics in this paper consistent with the leaderboard metric of the competition.

\subsection{Implementation details}
Our final model is the ensemble of a model incorporating extra-data mapping and the other model utilizing  contrastive loss. The first model is trained with crop size 896 on 6 GPUs with one image per GPU for 60k iterations, and the second model is trained with crop size 480 on 2 GPUs with 2 images per GPU for 40k iterations. We adopt AdanW optimizer in our training process with a learning rate of 2e-5 and a linear warmup of 1500 iters with ratio 1e-6. For data augmentation, random horizontal flipping, random cropping, and random resizing with a ratio range [0.5, 2.0] are used. In the testing phase, random horizontal flipping and a sliding window are applied for test-time augmentation. The backbone is pre-trained on ImageNet, and the whole model is pre-trained on ADE20K and COCO-Stuff. The experiments in this work are conducted with the support of the CMCC (China Mobile Communications Corporation) Jiutian Deep Learning Platform.

\subsection{Ablations Studies}
\textbf{Baseline} We explored some Transformer-based networks with different backbones and conducted a series of experiments, and results are shown in Table~\ref{tab:backbone_results}. Mask2former framework showed better performance compared to Upernet. Besides, Mask2former with beit2-adapter as backbone shows extraordinary performance among all the settings. 

\begin{table}[h!]
  \begin{center}
    \begin{tabular}{c|c | c } 
      \hline
      \textbf{Method} & \textbf{Backbone} & \textbf{Test1 mIoU}\\
      \hline
      UperNet\cite{xiao2018unified} & FocalNet\cite{yang2022focal} & 47.35\\
      UperNet & Swin\cite{liu2021swin} & 48.2\\
      UperNet & InternImage-XL\cite{wang2023internimage} & 48.52 \\ 
      Mask2former\cite{cheng2021mask2former} & Eva\cite{fang2023eva} & 57.45\\
      
      Mask2former & Beitv2-Adapter & \textbf{58.49}\\      
     \hline
    \end{tabular}
   \caption{Experiments of different backbones.}
   \label{tab:backbone_results}
   \end{center}
\end{table}

\textbf{Multi-dataset model} 
In this model, we adopted some components on the baseline model and obtained improvements. Firstly, enlarging the crop size during the training phase can boost performance. The possible reason can be that the model obtains more detailed information and the model was pre-trained with a crop size of 896 with positional embedding, which is conflicting when the number of patches changes. Secondly, utilizing the the mapped extra dataset in the training process. This operation greatly increased diversity of data. The details of the extra-data mapping are in ~\ref{subsec: Multi-dataset training}. Finally, we set the dropout rate to 0.5 which greatly improved the generalization performance of the model. We added the above settings step by step and the results are shown in Table \ref{tab:Multi-dataset model}. We can see that enlarging the crop size brings about 1.8 points improvements, using extra data adds 0.5 points, and the dropout skill obtains about 0.6 points improvements.
\begin{table}[h!]
  \begin{center}
    \begin{tabular}{c|c|c|c}
      \hline
      \textbf{Crop Size} & \textbf{mapping} & \textbf{dropout} & \textbf{Test1 mIoU}\\
      \hline
      480$\times$480 & $\times$ & $\times$ & 58.49\\
      896$\times$896  & $\times$ & $\times$ & 60.30\\
      896$\times$896  &$\surd$ & $\times$ & 60.77\\   
      896$\times$896  &$\surd$ & 0.5  & \textbf{61.32}\\   
     \hline
    \end{tabular}
   \caption{Ablation experiments on multi-dataset model.}
   \label{tab:Multi-dataset model}
   \end{center}
\end{table}

\textbf{Contrastive-loss model} 
In this study, we aimed to improve the continuity of predicted results between adjacent frames in video sequences. We introduced a contrastive loss function into the baseline model to accomplish this.The experimental results, as presented in Table~\ref{tab:contrastive loss}, demonstrate that the performance of the final model was significantly enhanced by using the contrastive loss function. In particular, we found that using two frames of images as input for the loss function yielded the best performance. However, It should be noted that we could not conduct experiments on images with a crop size of 896 or include more adjacent frames due to memory limitations. As a result, the following experimental results were based on images with a crop size of 480 and only two consecutive frames. 
Overall, our findings suggest that the use of a contrastive loss function can greatly improve the performance of predicted results in video sequences.

\begin{table}[h!]
  \begin{center}
    \begin{tabular}{c|c}
      \hline
      \textbf{Methods} & \textbf{Test1 mIoU} \\
      \hline
      Baseline & 58.05\\
      $NCE_4$ & 60.0\\
      $NCE_2$ &  \textbf{60.95} \\
     \hline
    \end{tabular}
   \caption{Ablation experiments on contrastive loss. }
   \label{tab:contrastive loss}
   \end{center}
\end{table}

where $NCE_4$ and $NCE_2$ represent the input of the contrastive loss function as two or four consecutive images, respectively, in one video.

\textbf{Model aggregation} 
We conducted experiments on two model aggregation methods, including soft ensemble and voting. 

As for soft ensemble, we adopted two ways of ensemble. In the first way, 
we applied a weighted summation on the soft result of the two chosen models
\begin{equation}
P = \tau {P}_{1}+(1-\tau) {P}_{2}
\end{equation}
where $\tau$ is the ensemble coefficient.
The two model ensemble results are shown in Table \ref{tab:soft ensemble}. 
In the other way, we average all soft model results.

\begin{equation}
P = -\frac{1}{N}\sum_{i=1}^{N} \mathcal{P}_{i}
\end{equation}
where ${P}_{i}$ represents the $i$ $th$ soft result and $N$ is the total number of models. In our experiments, we chose two multi-dataset model separately with crop size 480 and 896 with best results, and a contrastive-loss model.

Voting follows the principle of majority rule. We randomly chose some models and obtained the results, and then for every pixel in the image,  we selected the highest number of votes as the final class. In this phase, we randomly chose several models with competitive performance.

All the methods mentioned above have achieved expressive results in the final phase. The results are shown in Table \ref{tab:model aggregation}

\begin{table}[h!]
  \begin{center}
    \begin{tabular}{c|c | c}
      \hline
      \textbf{$\tau$} &\textbf{$1-\tau$} & \textbf{Test2 mIoU} \\
      \hline
      1 & 0 & 64.92\\
      0 & 1 & 63.56\\
      0.1 & 0.9 & 65.31\\
      0.2 & 0.8 & 65.68\\
      0.3 & 0.7 & 65.87\\
      0.4 & 0.6 & \textbf{65.95}\\
      0.5 & 0.5 & 65.94\\
      0.6 & 0.4 & 65.64\\
      0.7 & 0.3 & 65.44\\
      0.8 & 0.2 & 65.12\\
      0.9 & 0.1 & 64.79\\
     \hline
    \end{tabular}
   \caption{Ablation experiments on the soft ensemble. We apply $\tau$ on the Contrastive-loss model and $1-\tau$ on the Multi-dataset model.}
   \label{tab:soft ensemble}
   \end{center}
\end{table}

\begin{table}[h!]
  \begin{center}
    \begin{tabular}{c|c}
      \hline
      \textbf{Methods} & \textbf{Test2 mIoU} \\
      \hline
      Voting & 65.93\\
      Soft ensemble(multi model) & 65.72\\
      Soft ensemble(two model) & \textbf{65.95}\\
     \hline
    \end{tabular}
   \caption{Ablation experiments on model aggregation. }
   \label{tab:model aggregation}
   \end{center}
\end{table}

\subsection{Comparisons}
Our solution achieves 65.95\% mIoU on the final testing set and obtains the 1st place on the VSPW challenge at CVPR 2023. The results are shown in Table \ref{tab:Comparisons}

\begin{table}[h!]
  \begin{center}
    \begin{tabular}{c|c}
      \hline
      \textbf{Team} & \textbf{Test2 mIoU} \\
      \hline
      Ours & \textbf{65.95}\\
      SiegeLion & 65.83\\
      csj & 64.84\\
      naicha & 64.75\\
      chenguanlin & 63.72\\  
     \hline
    \end{tabular}
   \caption{Comparisons with other methods on the final test set.}
   \label{tab:Comparisons}
   \end{center}
\end{table}

\section{Conclusion}{
In this study, we started by selecting a strong baseline model that is well-suited for the task of multi-class semantic segmentation. To improve the performance of the model, we mapped an additional dataset to the VSPW dataset and employed a contrastive loss function to constrain temporal variations in the data. Additionally, we leveraged an effective model aggregation method to further enhance the overall performance of the system. All of these techniques were combined to create a comprehensive solution that achieved first place in the VSPW challenge at the prestigious CVPR 2023 conference. Our results demonstrate the effectiveness and versatility of our approach for addressing multi-task semantic segmentation problems, thereby offering the potential for improved performance in a wide range of scenarios. The noteworthy point is that we have many interesting attempts, such as post-processing the model results with optical flow, separately training individual categories, and designing imbalanced weights to address class imbalance issues, but ultimately, their performance on the test set was not stable. Therefore, we did not use these strategies in the final results.
}
{\small
\bibliographystyle{ieee_fullname}
\bibliography{egbib}
}

\end{document}